\documentclass[sigconf]{acmart}
\usepackage[utf8]{inputenc}
\usepackage{newunicodechar}
\newunicodechar{̒}{} 
\renewcommand\footnotetextcopyrightpermission[1]{}

\usepackage{microtype} 
\usepackage{booktabs}  
\usepackage{xcolor}
\usepackage{colortbl}
\usepackage{pifont}
\usepackage{multirow}
\usepackage{wrapfig}
\usepackage{graphicx}
\usepackage{url}
\usepackage{booktabs}
\usepackage{tabularx}
\usepackage{caption}
\usepackage{tabularx} 
\usepackage{booktabs} 
\usepackage{colortbl} 
\usepackage{pifont} 
\usepackage{graphicx}
\usepackage{fancyhdr}

\usepackage{epigraph}
\usepackage{tcolorbox} 
\usepackage{enumitem}
\usepackage[utf8]{inputenc} 

\AtBeginDocument{%
  }

\acmISBN{978-1-4503-XXXX-X/2018/06}

\definecolor{skyblue}{RGB}{	70 ,130 ,180} 
\definecolor{lightblue}{RGB}{135 ,206 ,250}
\begin{document}

\title{EmoArt\includegraphics[height=1.3em]{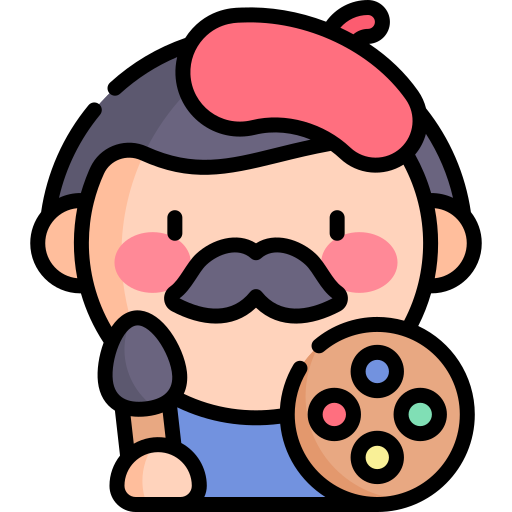}: A Multidimensional Dataset for Emotion-Aware Artistic Generation}

\author{Cheng Zhang}
\email{zhangcheng2122@mails.jlu.edu.cn}
\affiliation{%
  \institution{Jilin University}
  \city{Changchun}
  \country{China}
}

\author{Hongxia Xie}
\email{hongxiaxie@jlu.edu.cn}
\affiliation{%
  \institution{Jilin University}
  \city{Changchun}
  \country{China}
}

\author{Bin Wen}
\email{wenbin2122@mails.jlu.edu.cn}
\affiliation{%
  \institution{Jilin University}
  \city{Changchun}
  \country{China}
}

\author{Songhan Zuo}
\email{zuosh2122@mails.jlu.edu.cn}
\affiliation{%
  \institution{Jilin University}
  \city{Changchun}
  \country{China}
}

\author{Ruoxuan Zhang}
\email{zhangrx22@mails.jlu.edu.cn}
\affiliation{%
  \institution{Jilin University}
  \city{Changchun}
  \country{China}
}

\author{Wen-Huang Cheng}
\email{wenhuang@csie.ntu.edu.tw}
\affiliation{%
  \institution{National Taiwan University}
  \city{Taipei}
  \country{Taiwan}
}


\begin{abstract}
 With the rapid advancement of diffusion models, text-to-image generation has achieved significant progress in image resolution, detail fidelity, and semantic alignment, particularly with models like Stable Diffusion 3.5, Stable Diffusion XL, and FLUX.1. However, generating emotionally expressive and abstract artistic images remains a major challenge, largely due to the lack of large-scale, fine-grained emotional datasets. To address this gap, we present the EmoArt Dataset—one of the most comprehensive emotion-annotated art datasets to date. It contains 132,664 artworks across 56 painting styles (e.g., Impressionism, Expressionism, Abstract Art), offering rich stylistic and cultural diversity. Each image includes structured annotations: objective scene descriptions, five key visual attributes (brushwork, composition, color, line, light), binary arousal-valence labels, twelve emotion categories, and potential art therapy effects. Using EmoArt, we systematically evaluate popular text-to-image diffusion models for their ability to generate emotionally aligned images from text. Our work provides essential data and benchmarks for emotion-driven image synthesis and aims to advance fields such as affective computing, multimodal learning, and computational art, enabling applications in art therapy and creative design. The dataset and more details can be accessed via the following link:\href{https://zhiliangzhang.github.io/EmoArt-130k/}{\textcolor{skyblue}{https://zhiliangzhang.github.io/EmoArt-130k/}}
\end{abstract}
\begin{teaserfigure}
  \includegraphics[width=\textwidth]{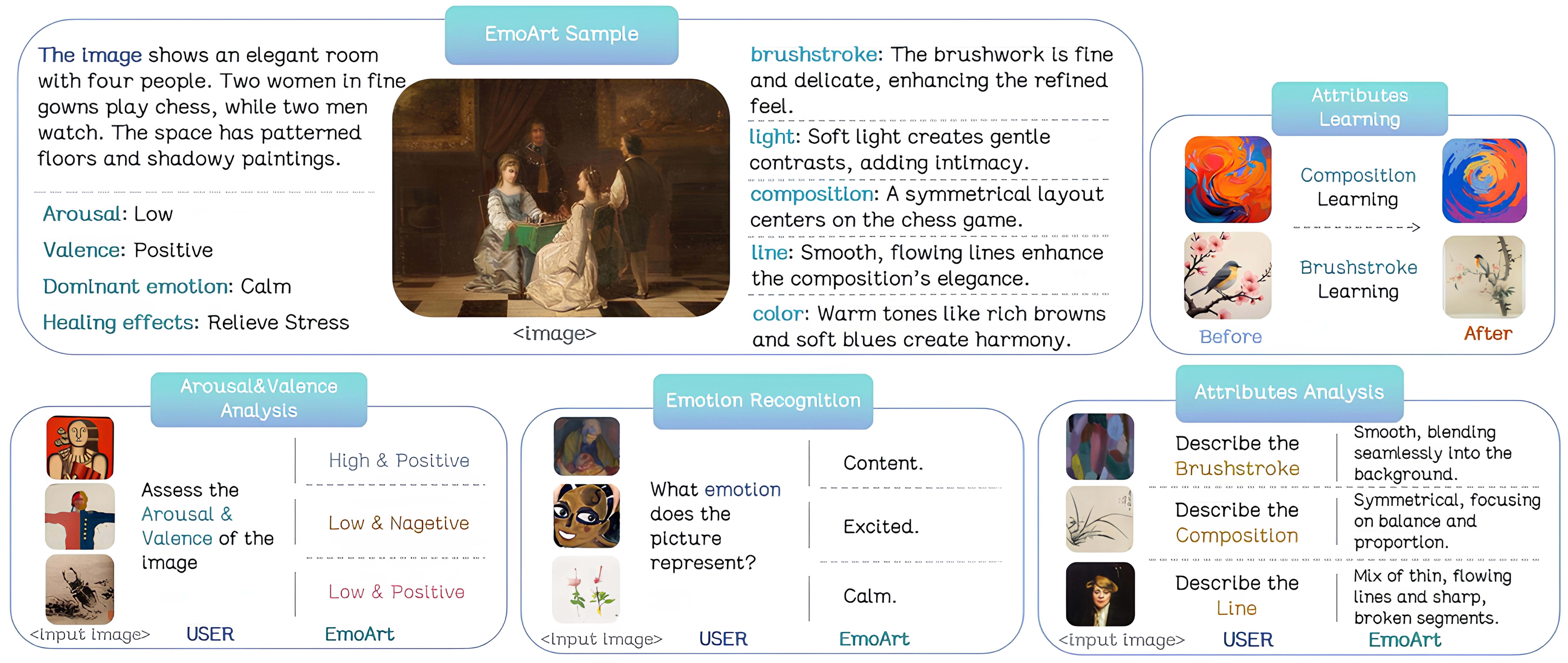}
  \caption{A Sample and Functional Modules of EmoArt. The figure illustrates the pipeline from input image to multi-level emotion and attribute analysis, as well as the system’s capability of learning visual styles such as brushstroke and composition.}
  \Description{Enjoying the baseball game from the third-base
  seats. Ichiro Suzuki preparing to bat.}
  \label{fig:teaser}
\end{teaserfigure}
\maketitle

\begin{table*}
\caption{Comparison of Emotion-related Datasets, R represents Recognition, G represents Generation.
}
\label{tab:datasets}
\renewcommand{\arraystretch}{1.2}
\begin{tabular}{lcccccccc}
\toprule
Dataset & Image Type & Label Source & Tasks & \#Image & Category & Valence\&Arousal & Attributes & Description \\
\midrule
IAPSa~\cite{mikels2005emotional} & Photo & Human & R & 395 & \textcolor[HTML]{228B22}{\ding{51}} & \textcolor[HTML]{228B22}{\ding{51}} & \textcolor[HTML]{CD5C5C}{\ding{55}} & \textcolor[HTML]{CD5C5C}{\ding{55}} \\
GAPED~\cite{dan2011geneva} & Photo & Human & R & 730 & \textcolor[HTML]{228B22}{\ding{51}} & \textcolor[HTML]{228B22}{\ding{51}} & \textcolor[HTML]{CD5C5C}{\ding{55}} & \textcolor[HTML]{CD5C5C}{\ding{55}} \\
ArtPhoto~\cite{machajdik2010affective} & Art & Human & R & 806 & \textcolor[HTML]{228B22}{\ding{51}} & \textcolor[HTML]{CD5C5C}{\ding{55}} & \textcolor[HTML]{CD5C5C}{\ding{55}} & \textcolor[HTML]{CD5C5C}{\ding{55}} \\
Emotion6~\cite{peng2015mixed} & Photo & Human & R & 1980 & \textcolor[HTML]{228B22}{\ding{51}} & \textcolor[HTML]{CD5C5C}{\ding{55}} & \textcolor[HTML]{CD5C5C}{\ding{55}} & \textcolor[HTML]{CD5C5C}{\ding{55}} \\
FI~\cite{you2016building} & Photo & Human & R & 23308 & \textcolor[HTML]{228B22}{\ding{51}} & \textcolor[HTML]{CD5C5C}{\ding{55}} & \textcolor[HTML]{CD5C5C}{\ding{55}} & \textcolor[HTML]{CD5C5C}{\ding{55}} \\
WEBEmo~\cite{panda2018contemplating} & Photo & Human & R & 268K & \textcolor[HTML]{228B22}{\ding{51}} & \textcolor[HTML]{CD5C5C}{\ding{55}} & \textcolor[HTML]{CD5C5C}{\ding{55}} & \textcolor[HTML]{CD5C5C}{\ding{55}} \\
Artemis~\cite{achlioptas2021artemis} & Art & Human & G\&R & 80K & \textcolor[HTML]{228B22}{\ding{51}} & \textcolor[HTML]{CD5C5C}{\ding{55}} & \textcolor[HTML]{CD5C5C}{\ding{55}} & \textcolor[HTML]{228B22}{\ding{51}} \\
EmoSet~\cite{yang2023emoset} & Photo/Art & Human\&LLM & G\&R & 3300K & \textcolor[HTML]{228B22}{\ding{51}} & \textcolor[HTML]{CD5C5C}{\ding{55}} & \textcolor[HTML]{228B22}{\ding{51}} & \textcolor[HTML]{CD5C5C}{\ding{55}} \\
FindingEmo~\cite{mertens2024findingemo} & Photo & Human & R & 25K & \textcolor[HTML]{228B22}{\ding{51}} & \textcolor[HTML]{228B22}{\ding{51}} & \textcolor[HTML]{CD5C5C}{\ding{55}} & \textcolor[HTML]{CD5C5C}{\ding{55}} \\
\rowcolor{lightblue!15}
EmoArt (Ours) & Art & Human\&LLM & G\&R & 130K & \textcolor[HTML]{228B22}{\ding{51}} & \textcolor[HTML]{228B22}{\ding{51}} & \textcolor[HTML]{228B22}{\ding{51}} & \textcolor[HTML]{228B22}{\ding{51}} \\
\bottomrule
\label{tab:datasets}
\end{tabular}
\end{table*}

\section{Introduction}
\begin{center}
    \vspace*{2em}
    \emph{``The purpose of art is washing the dust of daily life off our souls.''}
\end{center}
\begin{flushright}
    -- Pablo Picasso
\end{flushright}

The rapid development of AI-generated content (AIGC), especially diffusion-based text-to-image models like the Stable Diffusion series~\cite{sd21}, DALL·E~\cite{dall}, and Imagen~\cite{imagen}, has enabled realistic and semantically rich image synthesis. However, effectively conveying complex emotional expression remains a major challenge~\cite{you2015robust}.

Emotional intent is often faint, subjective, and context-dependent, making it difficult for generative models to interpret and reproduce with fidelity. Although \textbf{real-world image} synthesis has advanced, generating \textbf{artistic images} that convey complex emotions and deep affective meaning remains underexplored but essential, as art uniquely expresses complex emotions, culture, and therapeutic value beyond ordinary photos~\cite{artbook}. 

Existing emotion datasets such as AffectNet~\cite{mollahosseini2017affectnet}, EmoSet~\cite{yang2023emoset}, and ArtEmis~\cite{achlioptas2021artemis} are constrained by limited visual diversity, inconsistent labels, or insufficient support for multimodal emotion grounding.
To fill this gap, we introduce \textbf{\textit{EmoArt}}, a large-scale, multidimensional dataset designed to support both emotion understanding and generation in the artistic domain. EmoArt contains \textbf{132,664 paintings} spanning \textbf{56 stylistic genres} across a wide range of historical and cultural contexts, collected from The Met, WikiArt, and Europeana. Each image is annotated along three complementary dimensions: (1) \textit{content descriptions}, (2) \textit{visual attributes}, and (3) \textit{emotional and therapeutic effects}.


Annotations are generated via a GPT–4o–assisted pipeline with human verification, ensuring high consistency. We further benchmark leading diffusion models on emotional alignment and visual coherence. EmoArt aims to advance affective computing and computational art, while supporting emotion-aware and well-being–oriented applications.


Our contributions are summarized as follows:
\begin{itemize}[leftmargin=1em, itemsep=0pt, topsep=2pt]
  \item We introduce \textbf{EmoArt}, a large-scale, richly annotated dataset for emotion-aware image analysis and generation, covering 132,664 artistic images across 56 styles and three emotion-relevant dimensions.
  \item We benchmark state-of-the-art diffusion models on emotional alignment, validating EmoArt as a robust testbed for affective AIGC research.
\end{itemize}

\section{Comparison with Existing Emotion Datasets}



Existing emotion datasets in computer vision and affective computing include early works like \textit{ArtPhoto}~\cite{machajdik2010affective} and \textit{AbstractPhoto}~\cite{machajdik2010affective} focusing on artistic images with discrete labels; \textit{VSO}~\cite{borth2013large} and \textit{Twitter I/II}~\cite{you2015robust,you2016building} for social media sentiment; \textit{Emotion6}~\cite{peng2015mixed}, \textit{FI}~\cite{you2016building}, \textit{T4SA}~\cite{vadicamo2017cross}, and \textit{WEBEmo}~\cite{panda2018contemplating} covering diverse online images; and \textit{ArtEmis}~\cite{achlioptas2021artemis}, \textit{EmoSet}~\cite{yang2023emoset}, and \textit{FindingEmo}~\cite{mertens2024findingemo} providing large-scale and complex emotion annotations.

As shown in Table~\ref{tab:datasets}, existing datasets mostly target real-world photos, have limited or coarse annotations, or low image quality. In contrast, our \textbf{EmoArt} dataset offers large-scale, rich, and structured annotations designed specifically for emotionally-aware image generation.

\section{Construction of EmoArt}

\subsection{Data Collection and Filtering}

\begin{figure}[h]
  \centering
  \includegraphics[width=\linewidth]{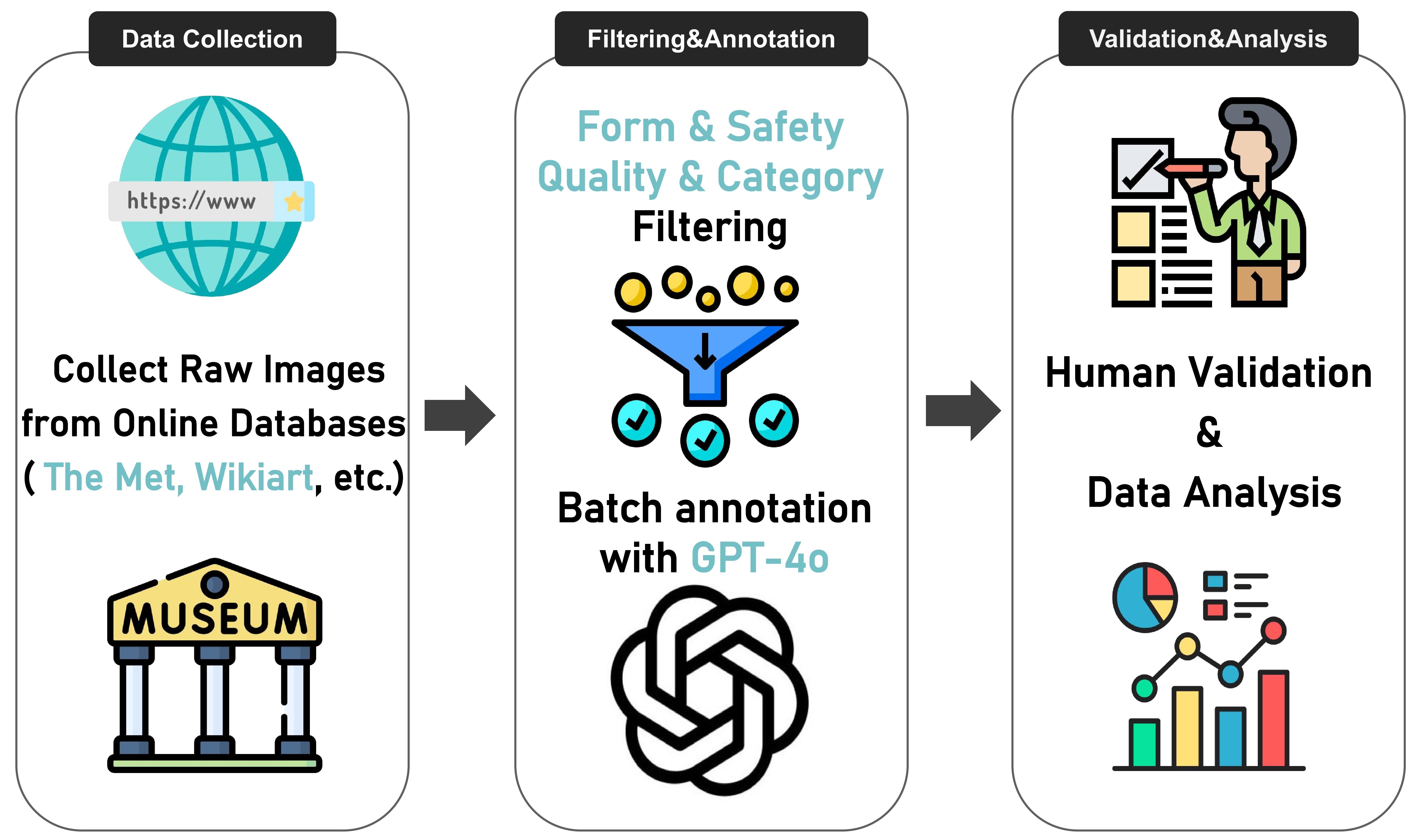}
  \caption{Construction pipeline of the EmoArt dataset.}
  \Description{A woman and a girl in white dresses sit in an open car.}
  \label{fig:pipeline}
\end{figure}

To construct the EmoArt dataset(see Figure~\ref{fig:pipeline}), we collected over 200,000 artworks representing more than 150 artistic styles from publicly accessible sources. These include \href{https://www.wikiart.org/}{WikiArt}, \href{https://www.metmuseum.org}{the Metropolitan Museum of Art}, \href{https://asia.si.edu/}{the National Museum of Asian Art}, \href{https://ukiyo-e.org/}{Japanese Print Search and Database}, \href{https://www.npm.gov.tw/}{the National Palace Museum (Taiwan)}, and \href{https://www.museum.go.kr/}{the National Museum of Korea}. The dataset covers both Western and non-Western traditions, ensuring broad regional and stylistic diversity. All works were sourced from public domain or open-access platforms, allowing for legal use and academic reproducibility.

To guarantee the dataset’s quality, representativeness, and ethical usability, we applied four rigorous filtering steps:

\begin{itemize}[leftmargin=1em]  
\item {\texttt{\bfseries Art Form Filtering}}: Retained only paintings; excluded non-painting media such as sculpture, crafts, prints, and photography to focus on emotional expression in painted works.
\item {\texttt{\bfseries Content Safety Filtering}}: Combined automated image classification with manual review to remove NSFW (Not Safe For Work) or explicit content, including some kitsch or overly suggestive artworks.
\item {\texttt{\bfseries Image Quality Filtering}}: Discarded images below 300×300 pixels or with visible compression artifacts, occlusions, or watermarks to ensure visual clarity and stable model training.
\item {\texttt{\bfseries Category Balance Filtering}}: Removed underrepresented styles (fewer than 400 samples) to maintain balanced distribution and ensure statistical robustness in analysis.
\end{itemize}
Through this systematic and thorough curation process, we obtained a legally usable and representative dataset of raw images.

\subsection{Data Annotation}

To leverage recent advances in multimodal intelligence, we adopt \textbf{GPT-4o}~\cite{achiam2023gpt} as the core annotation engine for the \textbf{EmoArt} dataset. With cutting-edge image understanding and affective modeling, GPT-4o interprets artistic images and produces structured annotations across visual and emotional dimensions. It processes image and text jointly, providing refined insights into visual semantics and simulating human-like emotional responses.

We design standardized \textit{prompt templates} and implement a multi-round generation-verification pipeline to ensure annotation quality and consistency. Empirical comparisons with human-labeled samples confirm GPT-4o’s strong alignment, validating its scalability for annotation.

\begin{figure}[h]
  \centering
  \includegraphics[width=\linewidth]{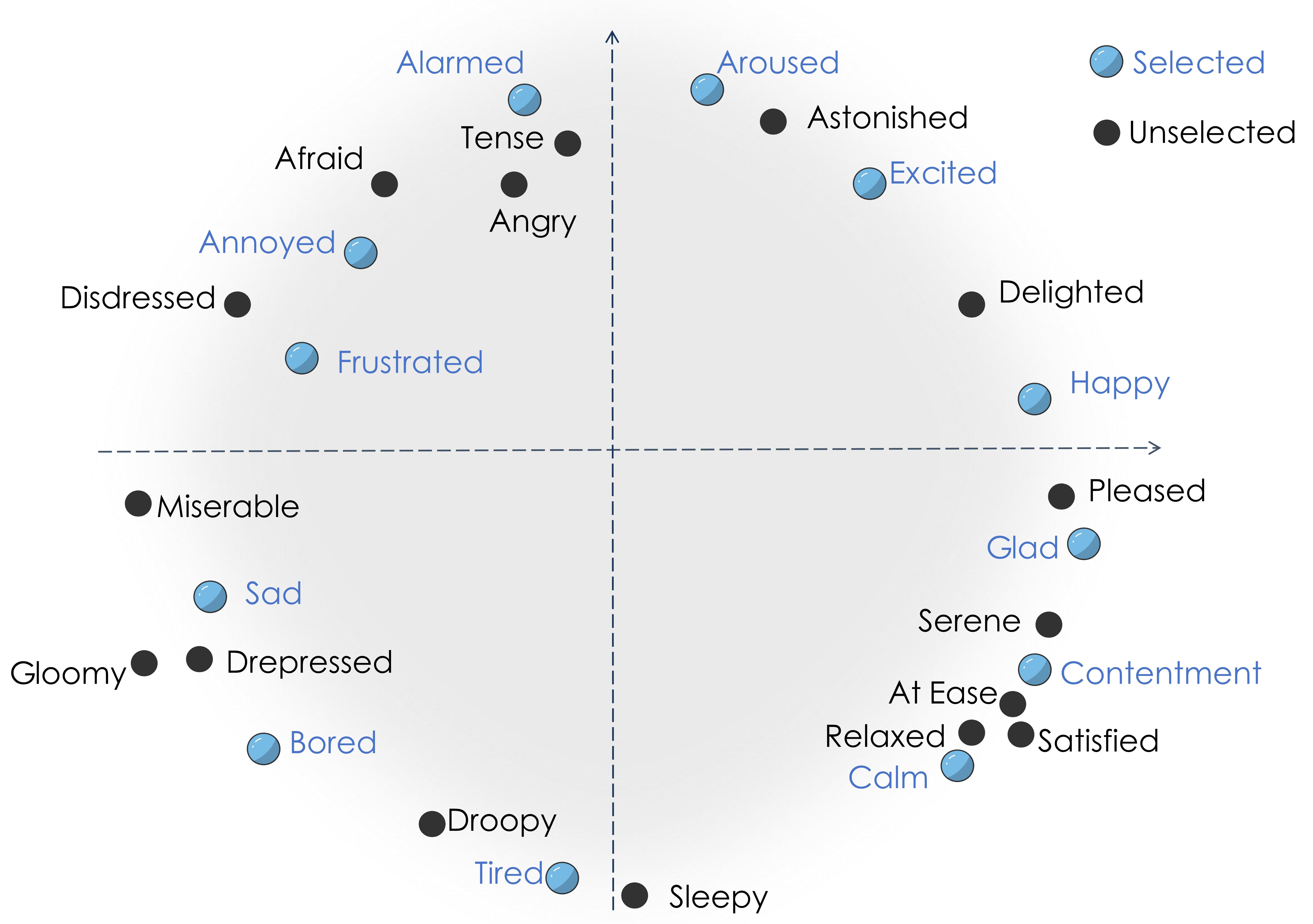}
  \caption{Distribution of 28 common emotions in the arousal-valence space and the selected 12 representative emotions.}
  \label{fig:russell}
\end{figure}

Compared to existing datasets, EmoArt adopts a more \textbf{hierarchical and multi-dimensional annotation framework}, capturing a spectrum from objective visual content to subjective emotion. Each image includes five key components:

\begin{itemize}[leftmargin=1em]  
    \item \textbf{Content Description}: Emotionally aware and detailed narratives lay the foundation for cross-modal understanding. Unlike COCO’s short captions or ArtEmis’s subjective responses, EmoArt integrates emotional cues like genre and composition, averaging 35.6 words per description—longer than ArtEmis~\cite{achlioptas2021artemis} (15.8) and COCO~\cite{chen2015microsoft} (10.5).
    
    \item \textbf{Visual Attributes}: Art’s emotional impact often comes from form. Drawing on art psychology, we annotate five structured features:
    \begin{itemize}
        \item \textbf{Brushwork}: Stroke thickness and rhythm;
        \item \textbf{Composition}: Symmetry and direction;
        \item \textbf{Color}: Hues and saturation;
        \item \textbf{Line}: Curvature and intensity;
        \item \textbf{Light}: Contrast and atmosphere.
    \end{itemize}
    This is the first emotion-centered dataset to annotate these attributes structurally.

    \item \textbf{Valence and Arousal}: Using Russell’s affect model~\cite{russell1980circumplex}, each image is labeled as:
    \begin{itemize}
        \item \textbf{Arousal (High / Low)}: Emotional intensity;
        \item \textbf{Valence (Positive / Negative)}: Pleasantness.
    \end{itemize}
    Binary labels improve clarity. For instance, a tranquil landscape may be “Low Arousal + Positive Valence,” while an intense scene may be “High Arousal + Negative Valence.”

    \item \textbf{Domain Emotion}: We select 12 representative emotions from 28, evenly covering the arousal-valence space:
    \begin{itemize}
        \item \textbf{Positive + High Arousal}: aroused, excited, happy;
        \item \textbf{Negative + High Arousal}: alarmed, annoyed, frustrated;
        \item \textbf{Negative + Low Arousal}: sad, bored, tired;
        \item \textbf{Positive + Low Arousal}: content, calm, glad.
    \end{itemize}
    GPT-4o assigns the most fitting label based on visual and emotional cues. Figure~\ref{fig:russell} illustrates the distribution of the selected 12 emotion categories.

    \item \textbf{Artistic Therapeutic Potential}: Inspired by art therapy~\cite{case2014handbook}, this label captures the artwork’s potential benefits—e.g., calming or uplifting—based on visual elements and emotional tone. For example, to relieve stress, a painting resembling Van Gogh’s \textit{Starry Night} could be generated.

\end{itemize}

This structured five-part annotation enables EmoArt to model the mapping from \textit{visual form} to \textit{emotional perception} to \textit{language generation} with high fidelity.

\begin{table*}[htbp]
  \centering
  \caption{Annotation Agreement Metrics across Annotation Sections.}
  \label{tab:agreement}
  \renewcommand{\arraystretch}{1.2}
  \begin{tabular}{lccccccc}
    \toprule
    \raisebox{0.6ex}{\textbf{Section}} & \textbf{\shortstack{True\\Proportion}} & \textbf{\shortstack{False\\Proportion}} & \textbf{\shortstack{Percent\\Agreement}} & \textbf{\shortstack{Positive\\Agreement}} & \textbf{\shortstack{Gwet's\\AC1}} & \textbf{\shortstack{McNemar\\p-value}} & \textbf{\shortstack{Sample\\Size}} \\
    \midrule
    Description & 98.01\% & 1.99\% & 94.25\% & 96.83\% & 0.928 & 0.38 & 5,922 \\
    Visual Attributes & 98.56\% & 1.44\% & 95.25\% & 97.87\% & 0.944 & 0.29 & 5,922 \\
    Emotion & 91.47\% & 8.53\% & 85.25\% & 90.14\% & 0.785 & 0.23 & 5,922 \\
    \bottomrule
  \end{tabular}
\end{table*}




\subsection{Human Validation}

We conducted a large-scale human validation on 5,600 images from the \textit{EmoArt} dataset, sampled across 56 artistic styles. Ten trained annotators independently assessed each image along three dimensions: \textbf{Description}, \textbf{Visual Attributes}, and \textbf{Emotion}.

As shown in Table~\ref{tab:agreement}, GPT-4o annotations showed high agreement with human labels: 98.01\% (Description), 98.56\% (Visual Attributes), and 91.47\% (Emotion), indicating strong alignment even in subjective categories.

We also measured inter-annotator reliability using standard metrics: overall/positive agreement, Gwet’s AC1, and McNemar’s test. All metrics confirmed high consistency (agreement $>$85\%, AC1 $>$0.75, $p$-values $>$ 0.05).

These results validate GPT-4o’s annotation quality and confirm \textit{EmoArt} as a reliable benchmark for emotion-aware image generation.

\section{Data Analysis}

\subsection{Distribution Analysis}
\begin{figure}[h]
  \centering
  \includegraphics[width=\linewidth]{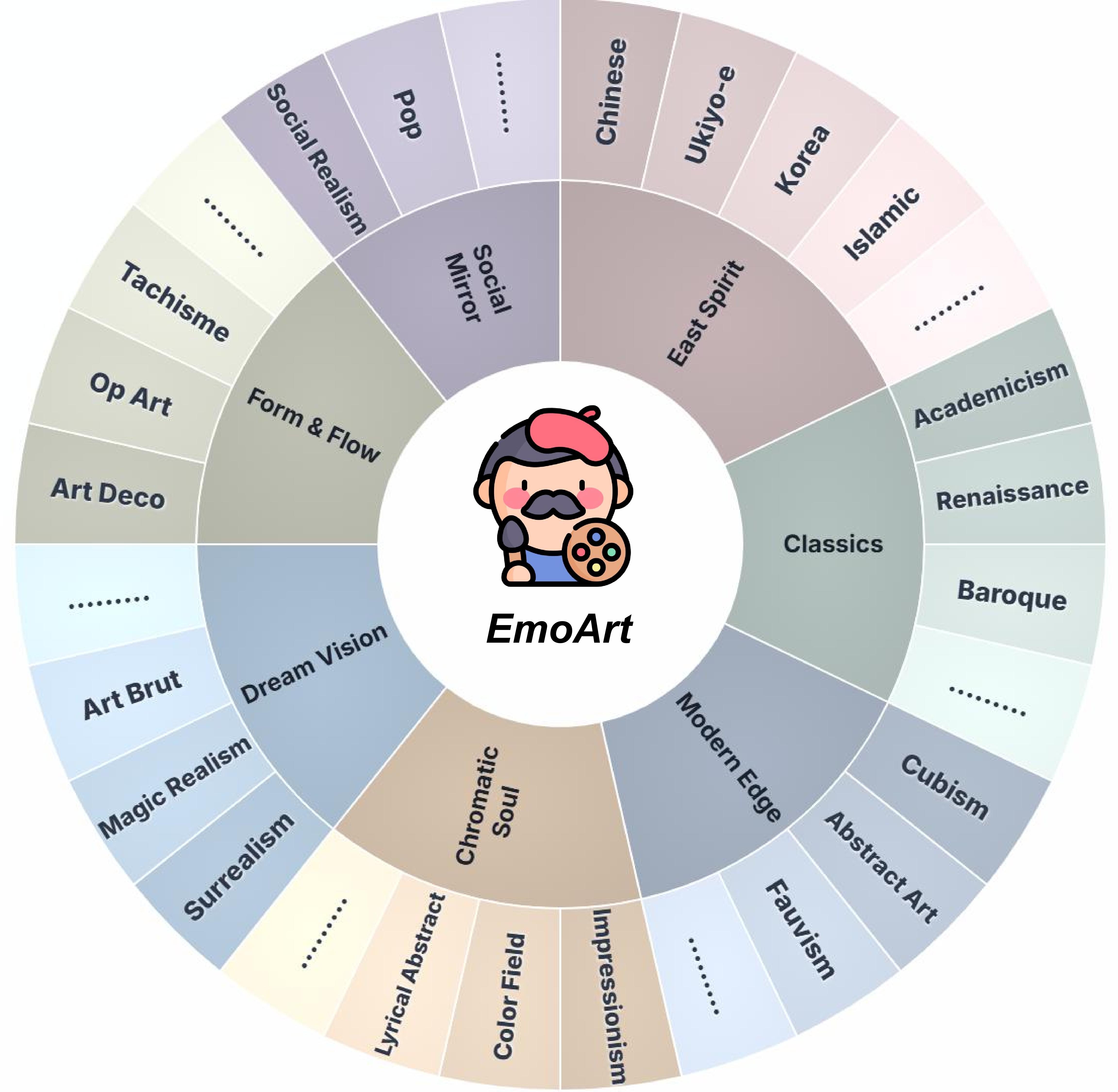}
  \caption{Representative art categories in the dataset: the inner ring shows the major categories, and the outer ring shows the specific subcategories.}
  \label{fig:sunburst}
\end{figure}

\begin{figure}[h]
  \Description{"Number of artworks in the top 10 artistic styles"}
  \centering
  \includegraphics[width=\linewidth]{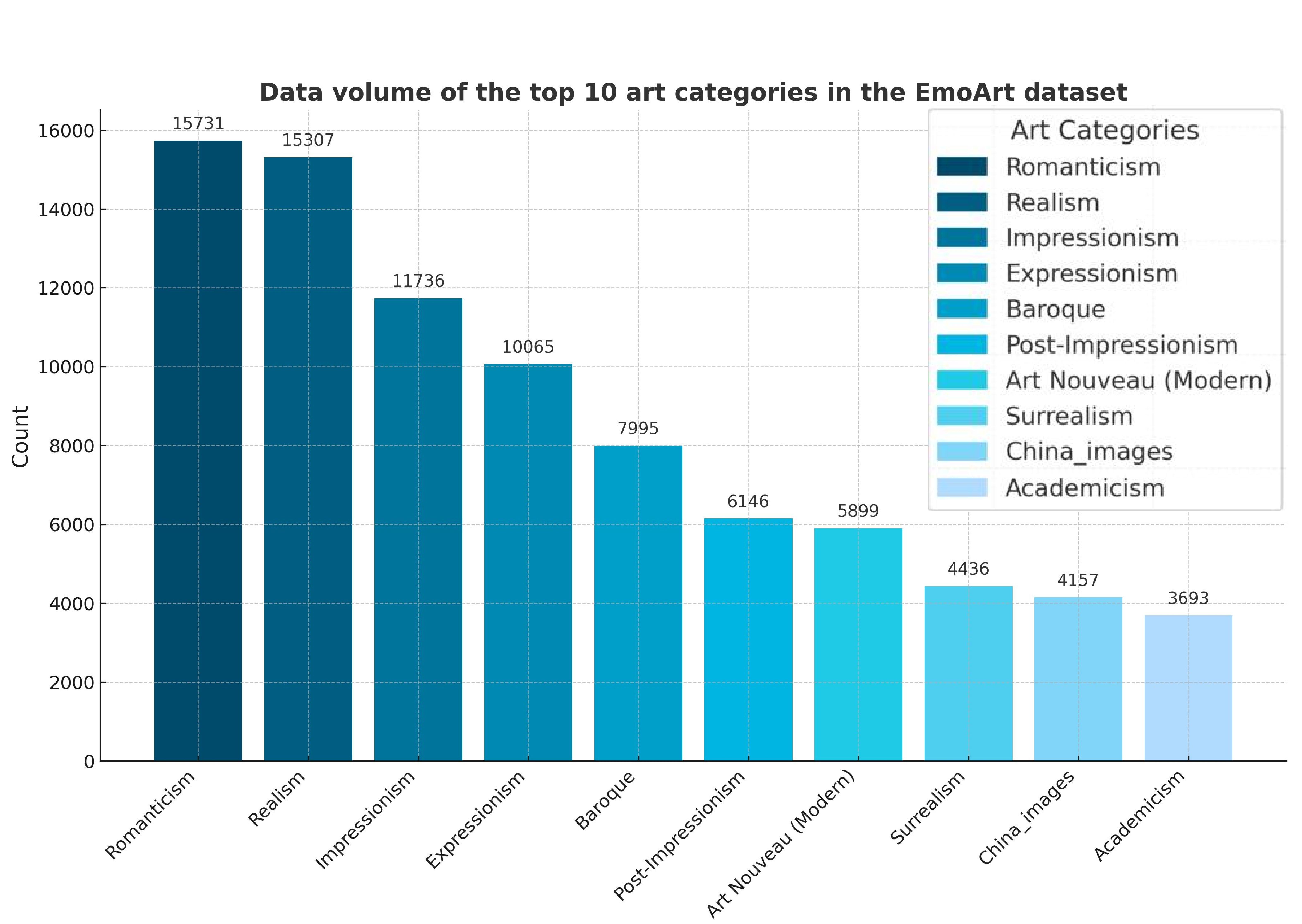}
  \caption{Data volume of the top 10 art categories in the EmoArt dataset.
}
  \Description{"Number of artworks in the top 10 artistic styles"}
  \label{fig:top10}
\end{figure}

The \textbf{EmoArt dataset} contains \textbf{132,664 samples} spanning \textbf{56 painting styles} across diverse historical periods and cultural contexts, from Early Renaissance to Neo-Pop Art. These styles are grouped into \textbf{seven thematic domains}—\textit{Classics}, \textit{Modern Edge}, \textit{East Spirit}, \textit{Chromatic Soul}, \textit{Dream Visions}, \textit{Form \& Flow}, and \textit{Social Mirror}—each reflecting distinct aesthetic and emotional characteristics. Representative samples are shown in Figure~\ref{fig:sunburst}, while Figure~\ref{fig:top10} illustrates the top 10 style categories by image count.

Emotion statistics indicate a strong tendency toward \textbf{positive valence (87.93\%)} and \textbf{low arousal (76.41\%)}, suggesting that most artworks evoke \textbf{pleasant and calming emotions}. The dominant labels are \textit{Calm} (55.95\%), \textit{Excited} (15.50\%), and \textit{Contentment} (15.35\%). Notably, \textbf{low arousal + positive valence} samples constitute \textbf{71.33\%} of the dataset, highlighting the prevalence of soothing and uplifting affect in artistic expression. High-arousal or negative emotions like \textit{Alarmed} (4.07\%) and \textit{Sad} (4.38\%) are comparatively rare.



\begin{table}[htbp]
  \centering
  \caption{Language Diversity Metrics Comparison Across Datasets.}
  \label{tab:language_metrics}
  \renewcommand{\arraystretch}{1.2}
  \begin{tabular}{lcccc}
    \toprule
    \raisebox{0.6ex}{\textbf{Dataset}} & 
    \textbf{\shortstack{Average\\TTR}} & 
    \textbf{\shortstack{Average\\MTLD}} & 
    \textbf{\shortstack{Average\\Entropy}} & 
    \textbf{\shortstack{Average\\Word Count}} \\
    \midrule
    Flickr30K & 0.9097 & 11.9349 & 3.3116 & 12.3392 \\
    ArtEmis & 0.9182 & 15.3065 & 3.6680 & 15.8919 \\
    COCO Cap. & 0.9065 & 10.1764 & 3.1567 & 10.4746 \\
    \rowcolor{lightblue!15}
    \textbf{EmoArt} & \textbf{0.9358} & \textbf{16.3396} & \textbf{3.6722} & \textbf{16.2184} \\
    \bottomrule
  \end{tabular}
  \label{tab:language}
\end{table}

\begin{table*}[htbp]
  \centering
  \caption{Image Generation Model Performance Metrics (Bruchstr. and Compo. stand for Brushstroke and Composition, respectively).}
  \label{tab:image_metrics}
  \renewcommand{\arraystretch}{1.2}
  \begin{tabular}{lcccccccccc}
    \toprule
    \raisebox{0.6ex}{\textbf{Model}} & 
    \begin{tabular}{@{}c@{}} \textbf{Brushstr.↑} \end{tabular} & 
    \begin{tabular}{@{}c@{}} \textbf{Color↑} \end{tabular} & 
    \begin{tabular}{@{}c@{}} \textbf{Compo.↑} \end{tabular} & 
    \begin{tabular}{@{}c@{}} \textbf{Light↑}  \end{tabular} & 
    \begin{tabular}{@{}c@{}} \textbf{Line↑} \end{tabular} & 
    \begin{tabular}{@{}c@{}} \textbf{Overall↑} \end{tabular} & 
    \begin{tabular}{@{}c@{}} \textbf{LPIPS↓} \end{tabular} & 
    \begin{tabular}{@{}c@{}} \textbf{SSIM↑} \end{tabular} & 
    \begin{tabular}{@{}c@{}} \textbf{PSNR↑} \end{tabular} & 
    \begin{tabular}{@{}c@{}} \textbf{FID↓} \end{tabular} \\
\midrule
    FLUX.1-dev & 0.6058 & 0.6703 & 0.6228 & \underline{0.6753} & 0.6216 & 0.6392 & \underline{0.6706} & \underline{0.2108} & 9.5708 & \textbf{21.2945} \\
    FLUX.1-schnell & 0.5875 & 0.6610 & 0.6263 & \textbf{0.6761} & 0.6250 & 0.6352 & 0.6947 & \textbf{0.2179} & 9.0199 & 38.1792 \\
    Playground & 0.6293 & \underline{0.6788} & 0.6247 & 0.6749 & 0.6354 & 0.6486 & 0.6715 & 0.1947 & \textbf{9.6673} & 42.5694 \\
    Pixart-sigma & \underline{0.6358} & 0.6746 & \underline{0.6342} & 0.6723 & \underline{0.6356} & \underline{0.6505} & 0.6754 & 0.1658 & 8.9910 & 36.2260 \\
    SDXL & 0.5939 & 0.6703 & 0.6257 & 0.6717 & 0.6311 & 0.6385 & 0.7110 & 0.1677 & 9.1273 & 61.9343 \\
    SD3.5 & 0.6211 & 0.6742 & 0.6324 & 0.6734 & 0.6317 & 0.6466 & 0.6991 & 0.1590 & 8.4539 & 37.9605 \\
    Openjourney & 0.6128 & 0.6380 & 0.6140 & 0.6620 & 0.6304 & 0.6314 & 0.7188 & 0.1480 & 9.0728 & 62.2185 \\
    \rowcolor{lightblue!15} \textbf{FLUX.1-dev-finetuned} & \textbf{0.6388} & \textbf{0.6974} & \textbf{0.6698} & 0.6542 & \textbf{0.6421} & \textbf{0.6604} & \textbf{0.6508} & 0.2102 & \underline{9.6596} & \underline{31.6510} \\
    \bottomrule
  \end{tabular}
  \label{tab:results}
\end{table*}

Emotional profiles vary significantly across styles and thematic domains. For example, \textit{Realism} (11.52\%) and \textit{Romanticism} (11.84\%), typically associated with the \textit{Classics} domain, are characterized by calm and peaceful emotions, with \textit{Calm} accounting for 64.73\% and 60.66\% of their respective samples, and low-positive emotional combinations dominating (84.55\% and 76.85\%). In contrast, styles such as \textit{Expressionism} (7.57\%) and \textit{Surrealism} (3.34\%), under \textit{Modern Edge} and \textit{Dream Visions} respectively, display heightened emotional intensity, with high arousal present in over 36\% of samples and notable proportions of negative affect (24.08\% and 28.99\%). These styles frequently evoke emotions like \textit{Alarmed} (5.20\% in Expressionism, 10.41\% in Surrealism) and \textit{Sad} (9.35\% and 8.57\%), reflecting their emphasis on inner turmoil and psychological depth.

Cultural variation is also pronounced. Within the \textit{East Spirit} domain, traditional Chinese painting (\textit{China\_images}, 3.13\%) overwhelmingly conveys calm and positive affective states, with 99.76\% of samples exhibiting low arousal and 99.95\% positive valence—\textit{Calm} alone accounts for 89.42\%. Similar patterns are found in \textit{Ukiyo-e} (86.13\% low arousal, 95.20\% positive valence) and \textit{Gongbi} (100\% low arousal and positive valence), reflecting Eastern aesthetic ideals of harmony, balance, and serenity.

Conversely, the \textit{Social Mirror} domain—including styles such as \textit{Social Realism} and \textit{Socialist Realism}—is marked by more intense and critical emotional content. These styles show significantly higher proportions of negative valence (42.53\% and 25.09\%, respectively) and elevated levels of \textit{Alarmed} responses (18.06\% and 9.82\%), consistent with their focus on social critique and depictions of human struggle.

\subsection{Linguistic analysis}

To evaluate the linguistic diversity and expressive complexity of the image description texts in the EmoArt dataset, we selected four commonly used quantitative metrics: TTR (Type-Token Ratio), MTLD (Mean Textual Lexical Diversity), lexical entropy (Entropy), and average word count, comparing the results with multiple mainstream visual-text datasets. 
As shown in Table~\ref{tab:language}, EmoArt consistently outperforms others across all metrics.
Specifically, EmoArt achieves a high TTR of 0.9358, an MTLD of 16.34, and a lexical entropy of 3.6722, indicating richer vocabulary and greater local variability. Its average word count of 16.22 further reflects more detailed and expressive descriptions than COCO Captions~\cite{chen2015microsoft} and Flickr30K~\cite{young2014image}.


These results confirm that EmoArt provides superior linguistic richness and information density, making it a strong foundation for tasks such as sentiment analysis, vision-language modeling, and text generation.

\section{Can AI Feel Art? Emotional Image Generation Benchmarks with EmoArt}

We conducted a comprehensive evaluation of several state-of-the-art text-to-image diffusion models on our proposed EmoArt dataset.

\subsection{Experimental Setup}

We established baselines using seven state-of-the-art diffusion models: \textbf{FLUX.1-dev~\cite{flux1ai2024}}, \textbf{FLUX.1-schnell~\cite{flux1ai2024}}, \textbf{SDXL~\cite{sdxl}}, \textbf{SD3.5~\cite{sd35}}, \textbf{PixArt-sigma~\cite{chen2024pixart}}, \textbf{Playground~\cite{li2024playground}}, and \textbf{Openjourney~\cite{openjourney}}.

To explore the effectiveness of EmoArt, we fine-tuned \textbf{FLUX.1-dev} using LoRA. The training used 50 curated paintings per artistic category, along with their \textit{Description}, \textit{Arousal}, and \textit{Valence} annotations. Fine-tuning was conducted on a single NVIDIA A100 GPU, and evaluation followed the same metrics as the baseline.



We evaluated the quality of generated images using a comprehensive set of metrics:

\begin{itemize}[leftmargin=1em]  
    \item \textbf{FID}: Assesses distributional similarity between generated and real images via Inception features. Lower is better.
    
    \item \textbf{SSIM}: Measures structural and perceptual similarity. Ranges from 0 to 1, with higher values indicating better visual similarity.
    
    \item \textbf{PSNR}: Quantifies reconstruction quality using mean squared error. Higher values imply lower distortion.
    
    \item \textbf{LPIPS}: Estimates perceptual similarity using deep features; lower scores indicate better alignment with human perception.
    
    \item \textbf{Attributes Alignment}: Our proposed metric evaluates semantic fidelity to five artistic attributes. We fine-tune MiniCPM-V-2.6 on EmoArt and compute similarity to ground-truth text in the CLIP embedding space.
\end{itemize}

\begin{figure*}[t]
  \centering
  \includegraphics[width=0.95\linewidth]{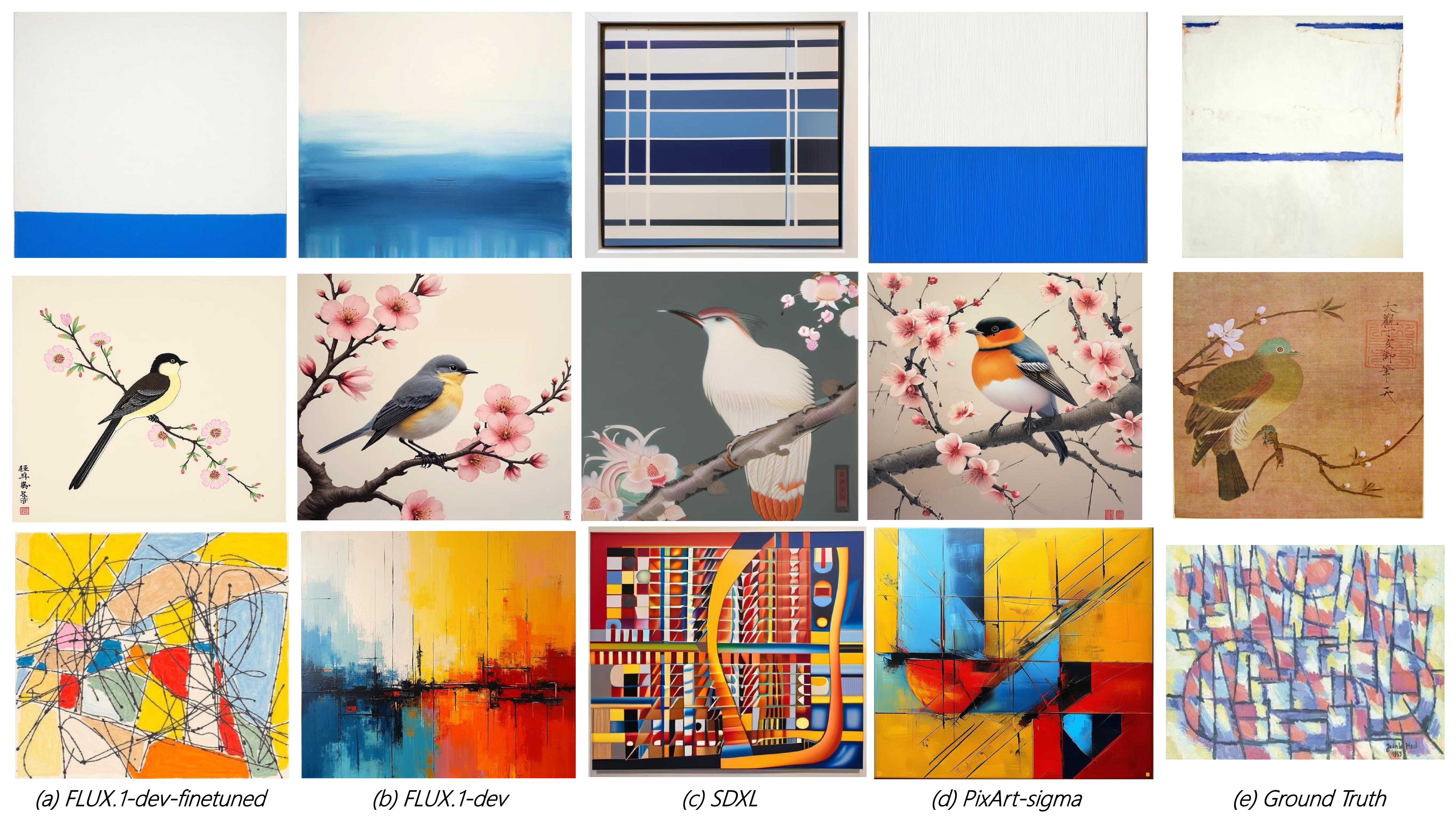}
  \caption{\textbf{Results from multiple text-to-image diffusion models.} The input prompt format is: \textit{Style + Arousal + Valence + Description}.}
  \label{fig:display}
\end{figure*}

\subsection{Quantitative Analysis}
\textbf{Question 1: Which model performs best in subjective evaluation metrics?} \\
\textbf{Answer:} \textit{FLUX.1-dev-finetuned} outperforms all other models across the majority of subjective evaluation metrics, demonstrating a clear advantage in perceived image quality.

As summarized in Table~\ref{tab:results}, \textit{FLUX.1-dev-finetuned}, trained with the proposed EmoArt dataset, achieves the highest scores in \textit{brushstroke} (0.6388), \textit{color} (0.6974), \textit{composition} (0.6698), \textit{line quality} (0.6421), and \textit{overall quality} (0.6604), reflecting strong alignment with human aesthetic judgments. This performance boost over its base model \textit{FLUX.1-dev} demonstrates the effectiveness of emotion-annotated fine-tuning in guiding stylistic and emotional rendering. The EmoArt dataset provides fine-grained supervision on visual elements and emotional intent, allowing the model to better internalize artistic patterns and generate images that resonate more deeply with viewers.

Interestingly, \textit{FLUX.1-schnell} slightly outperforms \textit{FLUX.1-dev} in \textit{light and shadow} (0.6761 vs. 0.6753), suggesting that its training configuration is particularly effective in capturing lighting dynamics, possibly due to better low-level feature representation.
Other models like \textit{Openjourney} and \textit{SDXL} exhibit moderate performance in \textit{overall quality} (0.6314 and 0.6385), but struggle with consistent brushstroke or compositional control, likely due to limited exposure to artistic styles during training. 

In general, these results indicate the importance of emotion-aware fine-tuning and structured artistic supervision in improving the subjective quality of generated images. Models trained with EmoArt not only achieve better emotional alignment but also exhibit enhanced stylistic authenticity.

\vspace{1em}
\vspace{1em}
\noindent\textbf{Question 2: How do the models perform in attribute alignment, and what insights does it offer for evaluation?} \\
\textbf{Answer:} \textit{FLUX.1-dev-finetuned} achieves the best results across most \textit{Attributes Alignment} metrics, revealing strong correlations between emotional dimensions and visual attributes, thereby supporting the validity of the \textit{EmoArt} annotations.

Specifically, \textit{FLUX.1-dev-finetuned} shows significant alignment between \textit{arousal} and \textit{valence} with core visual attributes such as \textit{brushstroke}, \textit{color}, \textit{light}, \textit{composition}, and \textit{line}. This indicates that the images generated by this model more effectively reflect the intended emotional content, reinforcing the scientific reliability of the \textit{EmoArt} framework.  

Although \textit{FLUX.1-dev-finetuned} does not achieve top performance in conventional evaluation metrics such as \textit{FID}, \textit{PSNR}, \textit{LPIPS}, and \textit{SSIM}, its near-optimal performance in \textit{Attributes Alignment} highlights the strength of the proposed framework. \textbf{These results suggest that traditional pixel-based metrics may not fully capture the perceptual and emotional quality of generated images, and that attribute evaluations can serve as a novel and complementary assessment perspective.}

\vspace{-0.5em}
\subsection{Qualitative Analysis}
\textbf{Question 3: How do the models differ in their ability to express emotion and artistic style in qualitative evaluations?} \\
\textbf{Answer:} In qualitative evaluations, \textit{FLUX.1-dev-finetuned} demonstrates a markedly superior capacity for emotional expression and stylistic fidelity across diverse artistic genres.

As illustrated in the first row of Figure~\ref{fig:display}, \textit{FLUX.1-dev-finetuned} effectively employs pure blocks of blue and white to evoke a calming atmosphere, faithfully capturing the essence of Color Field Painting. In contrast, \textit{FLUX.1-dev} and \textit{SDXL} generate images that lack stylistic clarity, exhibiting visual clutter and compositional inconsistency.

In the second row, which focuses on traditional Chinese painting, \textit{FLUX.1-dev-finetuned} adopts a minimalist, balanced composition and soft brushwork, conveying serenity and harmony aligned with East Asian aesthetics. Conversely, \textit{FLUX.1-dev}, \textit{SDXL}, and \textit{PixArt-sigma} rely on more vivid colors and intricate layouts, which diminish the subtle emotional tone intrinsic to this genre.

The third row evaluates depictions of high-arousal, anxious emotions. Here, \textit{FLUX.1-dev-finetuned} stands out with chaotic line work and asymmetric composition, effectively visualizing emotional intensity. While \textit{FLUX.1-dev}, \textit{SDXL}, and \textit{PixArt-sigma} also utilize non-equilibrium layouts, their outputs reveal template-like patterns and limited diversity, resulting in a less compelling emotional impact.

\section{Conclusion and License}

EmoArt offers 132{,}664 systematically annotated artworks spanning 56 diverse artistic styles, enabling fine-grained analysis and generation of emotionally expressive visual content. It serves as a valuable and comprehensive resource for affective computing, computational creativity, and multimodal learning across various research and application domains. The dataset is publicly available under the CC BY-NC 4.0 license at \href{https://huggingface.co/datasets/printblue/EmoArt-130k}{\textcolor{skyblue}{https://huggingface.co/datasets/printblue/EmoArt-130k}}.


\clearpage
\bibliographystyle{ACM-Reference-Format}
\bibliography{sample-base}
\clearpage


\end{document}